\documentclass[sigconf, nonacm]{acmart}

\AtBeginDocument{%
  }

\usepackage{multirow}
\usepackage{multicol}
\usepackage{makecell}
\usepackage{graphicx}
\usepackage{subcaption}
\usepackage{bm}
\usepackage{xcolor}
\usepackage{array}
\usepackage{url}
\usepackage{subcaption}
\usepackage{tabularx}
\usepackage{pgffor}

\begin{document}

\title{Rank-based No-reference Quality Assessment for Face Swapping}


\author{Xinghui Zhou}
\affiliation{%
  \institution{University of Science and Technology of China}
  \city{Hefei}
  \state{Anhui}
  \country{China}
}
\email{zhouxinghui@mail.ustc.edu.cn}

\author{Wenbo Zhou}
\affiliation{%
  \institution{University of Science and Technology of China}
  \city{Hefei}
  \state{Anhui}
  \country{China}
}
\email{welbeckz@ustc.edu.cn}

\author{Tianyi Wei}
\affiliation{%
  \institution{University of Science and Technology of China}
  \city{Hefei}
  \state{Anhui}
  \country{China}
}
\email{bestwty@mail.ustc.edu.cn}

\author{Shen Chen}
\affiliation{%
  \institution{Tencent AI Lab}
  \city{Shanghai}
  \country{China}
}
\email{kobeschen@tencent.com}

\author{Taiping Yao}
\affiliation{%
  \institution{Tencent Youtu Lab}
  \city{Shanghai}
  \country{China}
}
\email{taipingyao@tencent.com}

\author{Shouhong Ding}
\affiliation{%
  \institution{Tencent Youtu Lab}
  \city{Shanghai}
  \country{China}
}
\email{ericshding@tencent.com}

\author{Weiming Zhang}
\affiliation{%
  \institution{University of Science and Technology of China}
  \city{Hefei}
  \state{Anhui}
  \country{China}
}
\email{zhangwm@ustc.edu.cn}

\author{Nenghai Yu}
\affiliation{%
  \institution{University of Science and Technology of China}
  \city{Hefei}
  \state{Anhui}
  \country{China}
}
\email{ynh@ustc.edu.cn}

\begin{abstract}
Face swapping has become a prominent research area in computer vision and image processing due to rapid technological advancements. The metric of measuring the quality in most face swapping methods relies on several distances between the manipulated images and the source image, or the target image, i.e., there are suitable known reference face images. Therefore, there is still a gap in accurately assessing the quality of face interchange in reference-free scenarios. In this study, we present a novel no-reference image quality assessment (NR-IQA) method specifically designed for face swapping, addressing this issue by constructing a comprehensive large-scale dataset, implementing a method for ranking image quality based on multiple facial attributes, and incorporating a Siamese network based on interpretable qualitative comparisons. Our model demonstrates the state-of-the-art performance in the quality assessment of swapped faces, providing coarse- and fine-grained. Enhanced by this metric, an improved face-swapping model achieved a more advanced level with respect to expressions and poses. Extensive experiments confirm the superiority of our method over existing general no-reference image quality assessment metrics and the latest metric of facial image quality assessment, making it well suited for evaluating face swapping images in real-world scenarios.
\end{abstract}

\ccsdesc[500]{Security and privacy~Human and societal aspects of security and privacy}
\ccsdesc[500]{Computing methodologies~Computer vision tasks}
\ccsdesc[500]{Human-centred computing~Human computer interaction (HCI)}

\keywords{Deepfakes Detection, Face Swapping, Image Quality Assessment}



\settopmatter{printacmref=false} 
\renewcommand\footnotetextcopyrightpermission[1]{}
\maketitle
\begin{figure}
\centering
\includegraphics[width=\linewidth]{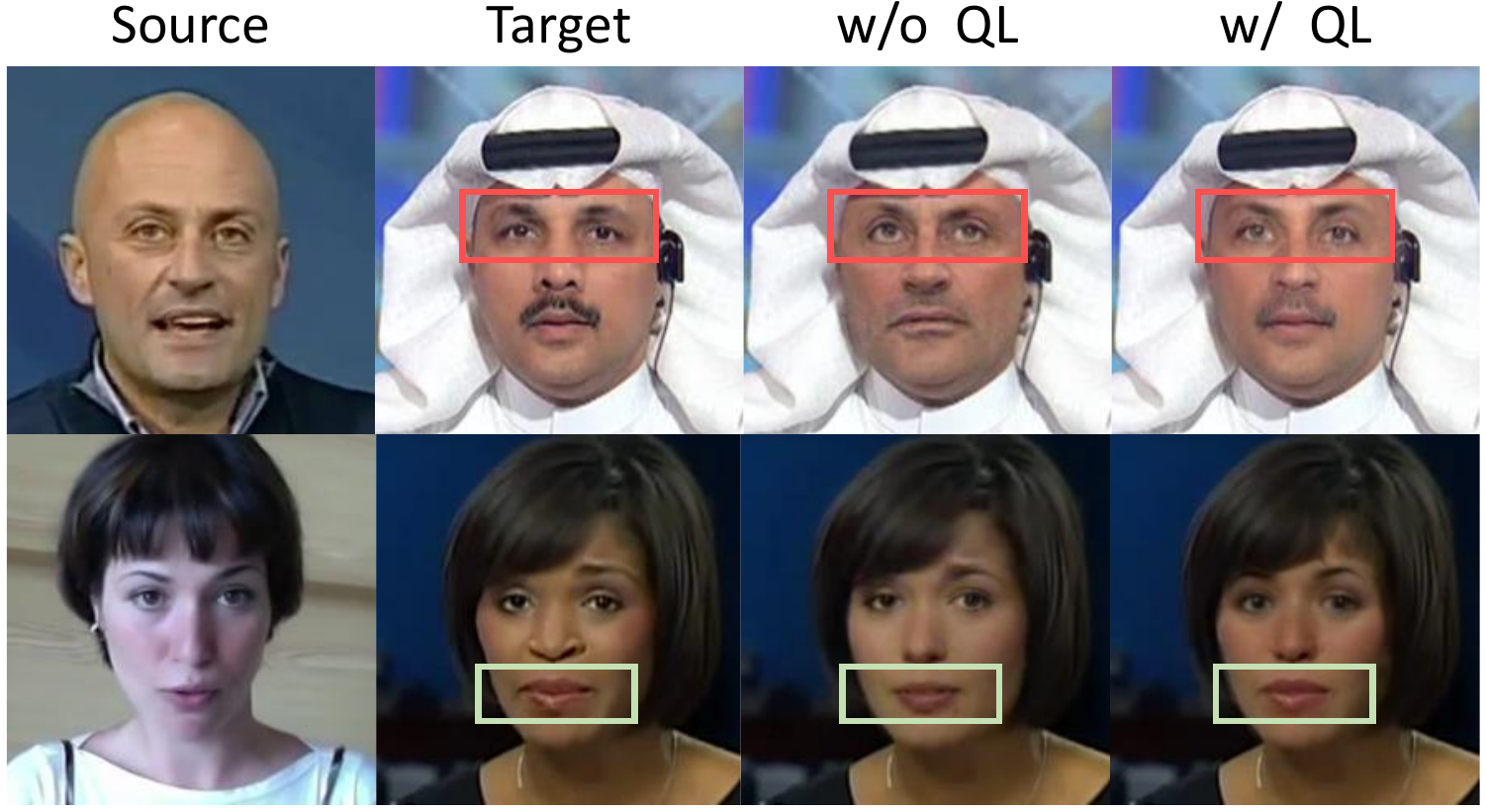}
\caption{Example of utilizing the proposed quality assessment metric as an additional loss constraint to improve the quality of face-swapping. The term `w/o QL' refers to training the swapping model with the original face-swapping loss function, while the term `w/ QL' refers to adding the proposed metric as an additional loss during training. The introduction of quality metric helps to maintain the gaze and expression of the target face while ensuring consistency of identity.}
\label{fig:teaser}
\vspace{-17pt}
\end{figure}

\section{Introduction}
With the continuous progress of generative models, face-swapping technology has become increasingly sophisticated, finding applications in fields such as movie production, virtual human creation, and privacy protection. The core of face swapping involves generating a forged face that preserves the identity of the source face while adopting other attributes from the target face.

Face swapping is increasing realism, yet it still lacks effective quality assessment, limiting its applications. Current full reference image quality assessment (FR-IQA) methods focus on distance errors related to facial expressions \cite{deng2019accurate, vemulapalli2019compact, feng2021learning}, pose \cite{chaudhuri2019joint, ruiz2018fine}, and shape \cite{deng2019accurate, sanyal2019learning}, using known target faces, which limits the applicability of the real world. Furthermore, diversity-based metrics such as Inception Score \cite{salimans2016improved}, Fréchet Inception Distance \cite{heusel2017gans} and SIFID \cite{shaham2019singan} fail to accurately assess face quality in forgeries. This highlights the need for new metrics that can evaluate face swaps more effectively in scenarios where target faces are unavailable, ensuring both realism and fidelity in generated images. 

Other specific methods include no-reference image quality assessment (NR-IQA) and face image quality assessment (FIQA). NR-IQA \cite{mittal2012no, talebi2018nima, ke2021musiq, golestaneh2022no, yang2022maniqa, zhang2023blind} focus on natural and transmission distortions such as noise, blockiness, blurring, and compression, while others \cite{gu2020giqa, guo2022assessing} focus on generative model distortions. FIQA \cite{hernandez2019faceqnet, terhorst2020ser, meng2021magface, ou2021sdd, jo2023ifqa} evaluates the quality of face image using face recognition models. Within FIQA, image quality is closely related to the proximity of facial feature embeddings to ideal class centroids in feature space. This subtle relationship is expressed as a positive correlation between image quality and intra-class distance and an inverse correlation with inter-class distance. However, these methods primarily assess quality based on identity, often ignoring non-identity attributes such as lighting and expression, which is crucial for a comprehensive understanding of face-swapping quality.

\begin{figure}[tb]
\centering
\includegraphics[width=\linewidth]{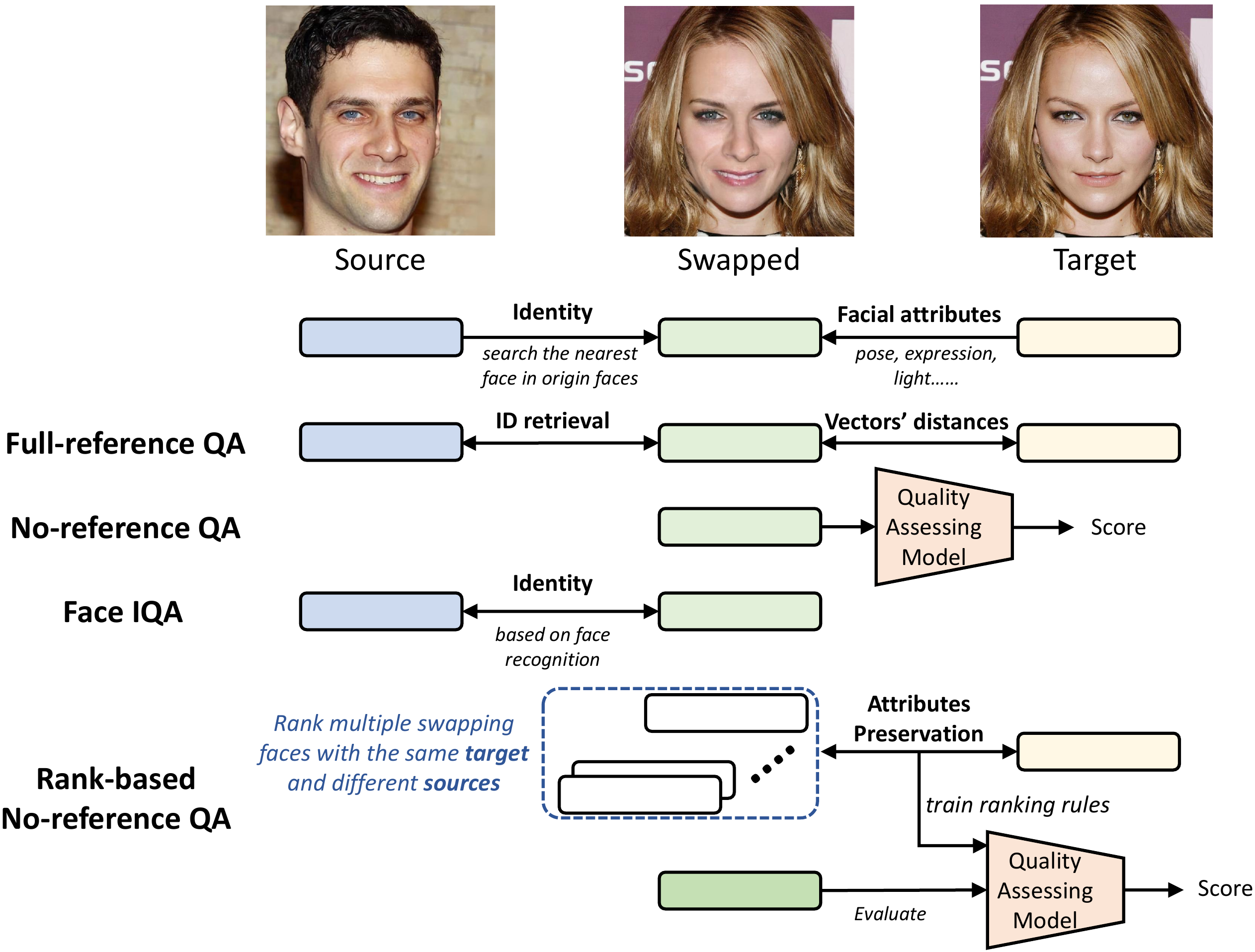}
\caption{Illustration of the three face-swapping related metrics and our metrics. (a) Full-reference quality assessment (FR-IQA) methods require the source and target faces to provide enough evaluations. (b) No-reference quality assessment (NR-IQA) methods require only exchanged faces, making it difficult to assess distortion due to specific identity embedding in face swapping. (c) Face IQA (FIQA) methods evaluates the identity of a face based on a face recognition task and cannot accurately evaluate synthetic images. (d) Our rank-based, no-reference quality assessment method ranks multiple swapped faces with the same target and different sources. By training consistency ranking rules that preserve attributes, we perform a new NR-IQA quality assessment.}
\label{fig:iqa_comparision}
\vspace{-10pt}
\end{figure}

To address these challenges, we introduce a rank-based, no-reference quality assessment method. Traditional no-reference approaches require extensive manual labeling, a costly process. Instead, we use image ranking to derive indirect labels. For accurate ranking labels that consider identity and attribute biases, we evaluate the consistency of the error between several attributes for various face-swap results with the same target face. These attributes cover lighting, pose, expression, and human perceptual similarity. The error consistency highlights uniform distortions relative to the source face's identity, guiding the generation of ranking pseudo-labels for image pairs. For getting enough face-swapping faces, we use the CelebAMask-HQ dataset \cite{lee2020maskgan}, have created more than $1$ million manipulated face images and generated more than $3.8$ million rank-based pseudo-labels with five forgery methods for our training dataset.

As an alternative to manual labeling, we generate reliable rank-based pseudo-labels to train our model. These ranking pseudolabels indicate the quality relationship between two manipulated faces in a pair. To accurately learn the rank, we used a Siamese network with two identical branches in a contrastive learning manner. During inference, only the parameters of one branch are used. The input of the model comprises two manipulated face images that feature the same target face. Training labels indicate the consistency ranking of image quality in this pair of face images (greater than or less than). To improve the robustness of the model, we employ a margin-aware ranking loss as the loss function for the Siamese network. 

Adequate human evaluation comparisons demonstrate that our approach responds well to subjective human perceptions. Furthermore, we utilize our metric to improve image quality in a state-of-the-art face-swapping architecture \cite{li2020advancing} to validate its effectiveness and relevance. As illustrated in Figure 1, by adding our metric as a loss function during training, the improved face swapping model can maintain the gaze and expression of the target face well, while ensuring consistency of identity.

The contributions of this paper are summarized as follows:

\begin{itemize}
\item We introduce a large-scale face swapping dataset that contains more than a million rank-based face swapping images produced by five different methods. The dataset contains rich ranking labels based on the consistency of face attribute vectors and perceptual similarity, labels correlated with face image quality to guide further assessment.
\item We propose a novel rank-based no-reference quality metric for face swapping images. Compared to other image quality assessment methods, our approach offers a comprehensive assessment that is consistent with human perception without reference images.
\item Extensive experiments on the human judgment consistency of visual realism assessment on deepfake faces demonstrate that our proposed method significantly outperforms others. We also add the metric as a loss in the training process to improve a known face-swapping model, resulting in lower attribute errors in quantitative comparisons.
\end{itemize} 

\section{Related Work} 
\subsection{Full-reference quality assessment}
To evaluate the quality of the swapped faces, a common method is to perform a full-reference quality assessment (FR-IQA) using existing sources and targets. For example, evaluating facial expression errors between swapped and target faces often involves 3D face reconstruction models such as MS-MFN \cite{chaudhuri2019joint}, D3DFR \cite{deng2019accurate}, and 3DDFAV2 \cite{guo2020towards}, which reconstruct expressive faces in 3D. Similarly, the pose of the head is assessed using HopeNet \cite{ruiz2018fine}, estimating the 3D pose by regressing the angles of yaw, pitch, and roll. For shape evaluation, works as RingNet \cite{RingNet_CVPR_2019} calculates the distance between the target and the swapped faces. However, 3D reconstruction and pose estimation models require high-quality human faces; lower quality can inaccurately capture each attribute's vector.

In recent years, more 3DMM-based face reconstruction models \cite{EMOCA_CVPR_2021, filntisis2022visual, wang2022faceverse, zielonka2022towards, khakhulin2022realistic, lin2023single} have offered improved accuracy in evaluating facial expressions on swapped faces. More outstanding head pose estimation models \cite{albiero2021img2pose, valle2020multi, hempel20226d} are proposed. Although D3DFR \cite{deng2019accurate} and HopeNet \cite{ruiz2018fine} remain the main assessment models to ensure the consistency of the environment for quantitative evaluation. In these methods, EMOCA \cite{EMOCA_CVPR_2021} achieves better expression reconstruction and is validated on AffectNet \cite{mollahosseini2017affectnet}.

Although these models facilitate an improved quantitative assessment of discrepancies between facial attributes, the acquisition of reliable target faces often poses a challenge, which restricts the practical deployment of FR-IQA in real-life situations. Consequently, for the quality evaluation of face-swapping, the development of a high-performing no-reference metric becomes essential.

\subsection{No-reference Quality Assessment}
One of the early works in No-Reference Quality Assessment (NR-IQA) was the Natural Scene Statistics (NSS) model proposed by \cite{simoncelli2001natural}. Over the years, several models and metrics based on mean subtracted contrast normalized (MSCN) have been proposed for NR-IQA, such as the Blind / Referenceless Image Spatial Quality Evaluator (BRISQUE) \cite{mittal2012no}. Other works, such as the Natural Image Quality Evaluator (NIQE) \cite{mittal2012making} and the Integrated Local NIQE (ILNIQE) \cite{zhang2015feature}, are influenced by image distortions. Recent research has explored various deep learning approaches to NR-IQA, including NIMA \cite{talebi2018nima} and PaQ-2-PiQ \cite{ying2020patches}. DBCNN \cite{zhang2018blind} proposes a deep bilinear model that predicts the quality of authentic and synthetic images. Recent work by MUSIQ \cite{ke2021musiq} and MANIQA \cite{yang2022maniqa} has achieved a no-reference prediction by incorporating vision transformers into their architectures. 

Some NR-IQA methods focus on assessing distortions in image generation, GIQA \cite{gu2020giqa} predicts quality based on the Gaussian mixture model and K-nearest neighbor of probability distributions on the generated images, and RISA \cite{guo2022assessing} based on the positive correlation between image quality and iterations.

While these methodologies have been widely applied in domains including face editing, image inpainting, and image super-resolution, NR-IQA and generative image assessment fall short in precisely detecting distortions arising from face swapping, particularly those affecting specific identity details. In essence, these models exhibit limitations in identifying distortions associated with the exchange of faces bearing distinctive identity features.

\subsection{Face Image Quality Assessment}
Face image quality assessment (FIQA) relies on the underlying face recognition model, such as FaceNet \cite{schroff2015facenet} and ArcFace \cite{deng2019accurate}. OFIQ \cite{kim2015face} used two factors, including visual quality and degree of mismatch between training and test images, to predict the quality of face images. SER-FIQ \cite{terhorst2020ser} proposed an unsupervised estimate of the quality of face images. They computed the average Euclidean distance of multiple embedded features from recognition models with different dropout patterns, as quality scores. \cite{xie2020inducing} proposed a Predictive Confidence Network (PCNet), where they applied pairwise regression loss to train a neural network from intra-class similarity for FIQA. MagFace \cite{meng2021magface} proposes an estimation method based on feature uncertainty to assess the quality of the face image. SDD-FIQA \cite{ou2021sdd} evaluates quality by measuring the distance between the distribution of intra-class similarity and the distribution of inter-class similarity of images. CR-FIQA \cite{boutros2023cr} assesses image quality based on class center angular similarity (CCAS) and close to the closest negative class center angular similarity (NNCAS). 

These methods assess the quality of images solely on the basis of single-identity information, overlooking attributes unrelated to identity, such as lighting and expression. Currently, constrained by the domain of training data, FIQA is often limited to evaluating faces in real-world scenarios. As a technique within deepfakes, face swapping exacerbates the challenge by introducing discrepancies between the distribution domains of synthesis and real data, rendering FIQA less effective in generalizing to face images with manipulated identity.

\begin{figure*}[htp]
\centering
\includegraphics[width=0.8\linewidth]{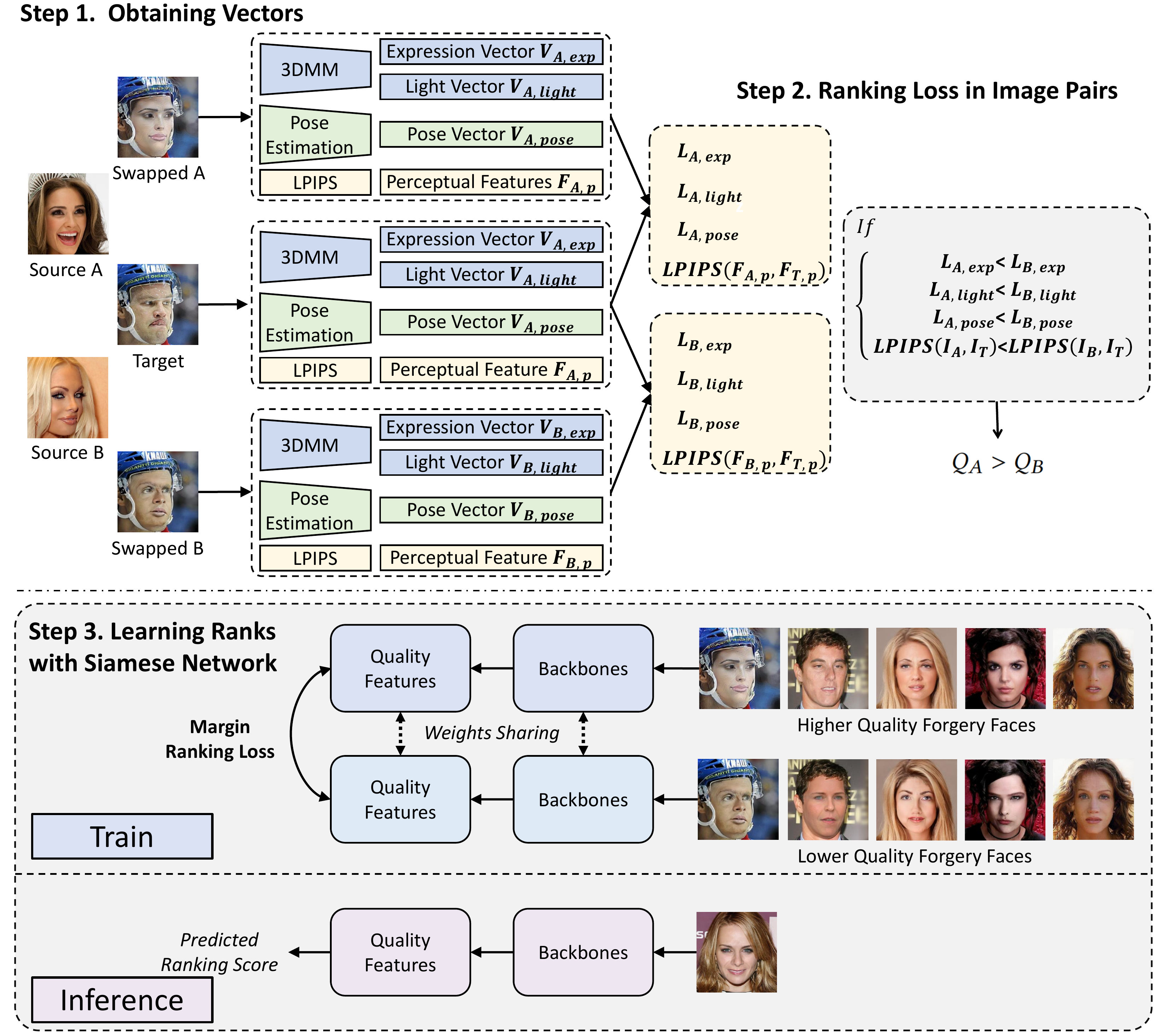}
\caption{The pipeline of rank-based labels generation. The following three components make up the generation of labels: 3D Face Reconstruction for Expression \& lighting Vector, Pose Vector Estimation and Label Generation.}
\label{fig:main}
\end{figure*}

\section{Rank-based No-reference Quality Assessment}
\subsection{Rank-based Labels Generation}
\textbf{Dataset Creation.}
To ensure an accurate and comprehensive prediction of image quality in face replacement, we propose building a large new database of facial forgeries. Specifically, we utilized high-resolution face images from the CelebAMask-HQ dataset \cite{lee2020maskgan} as the source and target images. We employ five advanced face-swapping methods: SimSwap \cite{chen2020simswap}, FaceShifter \cite{li2020advancing}, MegaFS \cite{zhu2021one}, HiRes \cite{xu2022high}, and InfoSwap \cite{gao2021information}. Each selected image was used as the target image and 10 other images were randomly selected as the source images for each forgery method. A total of $1,238,682$ images were generated and a general face detection algorithm \cite{deng2020retinaface} was used to remove images that did not recognize faces, excluding those generated incorrectly or without detectable faces.

The dataset we constructed adheres to the following two rules:

\textbf{1. Including self-swapped faces:} This setting allows the model to minimize interference with identity information when evaluating image quality. Additionally, the distortion present in self-swapped face images can be considered as a type of distortion introduced by the face-swapping model.

\textbf{2. Balanced data distribution:} The dataset contains low- and high-resolution images. Images with a resolution of 256x256 constitute 42\% of the total, those with a resolution of 512x512 account for 26\%, and images with a resolution of 1024x1024 represent 32\%. Furthermore, the number of faces generated by each of the five face-swapping methods is approximately equal. More details of the dataset can be found in the supplementary material.

To generate rank-based labels, we employed perceptual similarity and three vectors: expression, lighting, and pose.

\noindent\textbf{3D Face Reconstruction for Expression \& lighting Vectors.}
For the face region, we use an advanced model for 3D face reconstruction, EMOCA \cite{EMOCA_CVPR_2021}, which is better at reconstructing facial expressions while performing 3D face reconstruction on images. The facial model used in this model is FLAME \cite{li2017learning}, which includes face identity \(\bm{\beta} \in \mathbb{R}^{|\bm{\beta}|}\), the facial expression \(\bm{\psi} \in \mathbb{R}^{|\bm{\psi}|}\), and pose parameter \(\bm{\theta} \in \mathbb{R}^{3k + 3}\) for local and global rotation angles around the joints \(k\) (neck, chin, and both eyes, so \(k=4\)). Taking into account all parameters, FLAME \cite{li2017learning} outputs a face with \(n = 5023\) vertices of the mesh. Formally, therefore, the statistic FLAME face model is expressed as follows. 
\begin{equation}
M(\bm{\beta},\bm{\psi},\bm{\theta}) = W(T_P(\bm{\beta},\bm{\psi},\bm{\theta}),\mathbf{J}(\bm{\beta}), \bm{\theta}, \mathcal{W})
\end{equation}
which \(W(\mathbf{T}, \mathbf{J}, \bm{\theta}, \mathcal{W})\) is blend skinning function that rotates the vertices in \(\mathbf{T}\in\mathbb{R}^{3n}\) around joints \(\mathbf{J}\in\mathbb{R}^{3k}\), \(\mathcal{W}\in\mathbb{R}^{k\times n}\) is blendweights.

Further,
\begin{equation}
T_{P}(\bm{\beta}, \bm{\psi},\bm{\theta}) = \mathbf{T} + B_{S}(\bm{\beta};\mathcal{S}) + B_{P}(\bm{\beta};\mathcal{P}) + B_{E}(\bm{\beta};\mathcal{E})
\end{equation}
denotes the mean template \(\mathbf{T}\) in ``zero poses'' with added shape blendshapes \(B_{S}(\bm{\beta};\mathcal{S}):\mathbb{R}^{|\bm{\beta}|}\rightarrow\mathbb{R}^{3n}\), pose correctives \(B_{P}(\bm{\beta};\mathcal{P}): \mathbb{R}^{3k+3}\rightarrow\mathbb{R}^{3n}\), and expression blendshapes \(B_{E}(\bm{\beta};\mathcal{E}): \mathbb{R}^{|\bm{\psi}|}\rightarrow\mathbb{R}^{3n}\), with the learned face identity, pose, and expression base parameters \(\mathcal{S}\), \(\mathcal{P}\), \(\mathcal{E}\). 

In the 3D face reconstruction processing, EMOCA \cite{EMOCA_CVPR_2021} refers to the lighting model in DECA \cite{guo2020towards}, the shaded face image is computed as:
\begin{equation}
    B(\bm{\alpha}, \mathbf{L}, N_{uv})_{i,j} = A(\bm{\alpha})_{i,j} \odot \sum_{k=1}^{9}\mathbf{L}_{k}H_{k}(N_{i,j})
\end{equation}
where, \(A\) denotes albedo, \(N\) is surface normal, \(N_{uv}\) is FLAEM UV layout. \(B\) is shaded texture represented in UV coordinates and where \(B_{i,j}\in\mathbb{R}^{3}\), \(A_{i,j}\in\mathbb{R}^{3}\), and \(N_{i,j}\in\mathbb{R}^{3}\)3 denote the pixel \((i, j)\) in the UV coordinate system. The Spherical Harmonics basis and coefficients are defined as \(H_{k}:\mathbb{R}^{3}\rightarrow\mathbb{R}\) and \(\mathbf{L} = \begin{bmatrix}\mathbf{L}_1^T, & \cdots & ,\mathbf{L}_9^T \end{bmatrix}^T\), with \(\mathbf{L}_k\in\mathbb{R}^{3}\) and \(\odot\) denoting the Hadamard product.

We name the encoder for the output of the expression parameters as \(E_{c}\), it encodes the input image \(I_{i}\) for output several parameters: 
\begin{equation}
    E_{c}(I_{i}) =  (\bm{\beta}_{i}, \bm{\psi}_{i}, \bm{\theta}_{i}, \bm{\alpha}_{i}, \mathbf{L}_{i}, N_{uv,i})
\end{equation}
this set of parameters should reconstruct image \(I_{i}\) well. Formally, we minimise
\begin{equation}
    L_{sc} = L_{rec}(F_{i}, \mathcal{R}(M(\bm{\beta}_{i},\bm{\psi}_{i},\bm{\theta}_{i}), B(\bm{\alpha}_{i}, \mathbf{L}_{i}, N_{uv,i}), c_{i}))
\end{equation}
when \(L_{sc}\) tends to 0:  
\begin{equation}
    (\bm{\beta}_{i},\bm{\psi}_{i},\bm{\theta}_{i}, \bm{\alpha}_{i}, \mathbf{L}_{i}, N_{uv,i}) \rightarrow (\bm{\beta}_{gt},\bm{\psi}_{gt},\bm{\theta}_{gt}, \bm{\alpha}_{gt}, \mathbf{L}_{gt}, N_{uv,gt})
\end{equation}
where \(F_{i}\) means the identity embedding of the image \(I_{i}\), \(L_{sc}\)is shape consistency loss in 3D reconstruction. As \(L_{rec}\) converges to 0, \(\bm{\psi}_{i}\) and \(\mathbf{L}\) can make the rendered images look like real people. In the 3D reconstruction process of EMOCA, there are two different solutions: coarse and detailed. Different reconstruction methods can affect the loss \(L_{rec}\) is designed to be differently, and in this paper, we use the more accurate detail reconstruction settings.

In the 3D reconstruction process of EMOCA, we utilize the identity embedding \(F_{i}\) of an image \(I_{i}\) to generate realistic images that resemble the person's actual appearance. This is achieved by minimizing the reconstruction loss function \(L_{reconst}\), which produces increasingly accurate images as it approaches zero. 

\textbf{Pose Vector Estimation.}
Instead of directly learning the Euler angles of the face poses, we utilized a rotation matrix estimation model called 6DRepNet \cite{hempel20226d}. This approach overcomes the gimbal lock issue inherent in the Euler angle representation by directly predicting the rotation matrix. However, in the context of the Special Orthogonal Group \(SO(3)\), the rotation matrix must have a size of \(3 \times3\) and satisfy the orthogonality constraint. This constraint can be enforced through a Gram-Schmidt process or by finding the nearest optimal solution using SVD.

In the 6DRepNet model, we followed the approach introduced in \cite{zhou2019continuity} and performed a Gram-Schmidt mapping within the representation. This mapping involves discarding the last column vector of the rotation matrix, effectively reducing the \(3 \times 3\) matrix into a six-parameter rotation representation. This reduction has been confirmed to introduce smaller errors when directly regressing the pose vectors. By employing the 6DRepNet model with this modified rotation representation, we address the limitations associated with Euler angles and ensure a more accurate estimation of face poses. This reduces the \(3 \times3\) matrix into a six-parameter rotation representation, which introduces smaller errors when used for direct regression.
\begin{equation}
    \bm{\theta}_{i} = \begin{bmatrix}| & | & | \\ a_{1} & a_{2} & a_{3} \\ | & | & | \end{bmatrix}
\end{equation}
where \(\bm{\theta}_{i}\) belongs to   \(\mathbb{R}^{3k + 3}\). In this approach, for the sake of simplicity in calculations, we only consider the rotation angle of the rotation matrix around the eyes. Hence, \(k\) is set to 2.

After the Gram-Schmidt process:
\begin{equation}
    \bm{\theta}_{GS,i} = \begin{pmatrix}\begin{bmatrix}| & | & | \\ a_{1} & a_{2} & a_{3} \\ | & | & | \end{bmatrix}\end{pmatrix} = \begin{bmatrix}| & | \\ a_{1} & a_{2} \\ | & | \end{bmatrix}
\end{equation}

The predicted 6D representation matrix can then be mapped back to \(SO(3)\).
\begin{equation}
    \bm{\theta}_{6D,i} = \begin{pmatrix}\begin{bmatrix}| & | \\ a_{1} & a_{2} \\ | & | \end{bmatrix}\end{pmatrix} = \begin{bmatrix}| & | & | \\ b_{1} & b_{2} & b_{3} \\ | & | & | \end{bmatrix}
\end{equation}
Thus, the remaining column vector is simply determined by the cross product that ensures that the orthogonality constraint is satisfied for the resulting \(3 \times 3\) matrix.
\begin{equation}
\begin{aligned}
        b_{1} &= \frac{a_{1}}{\left\| a_{1} \right\|} \\
        b_{2} &= \frac{u_{2}}{\left\| u_{2} \right\|}, u_{2} = a_{2} - (b_{1} \cdot a_{2})b_{1} \\
        b_{3} &= b_{1} \times b_{2}
\end{aligned}
\end{equation}
As a result, pose estimating has predicted 6 parameters that are mapped into a \(3 \times 3\) rotation matrix in a subsequent transformation which at the same time also satisfies the orthogonality constraint.

\textbf{Label Generation.}
In face swapping, preserving similarity and facial attributes are two important aspects of image quality assessment (IQA). Using the target image as a reference, we can rank the swapped images on the basis of their attribute losses. To ensure that multiple attributes are evaluated independently, we propose an attribute-based ranking label for no-reference image quality assessment. This label is generated when the following conditions are satisfied: \(I_{b}\) retains more detail than \(I_{a}\), implying that \(I_{b}\) outperforms \(I_{a}\) in terms of attribute preservation.

The losses of the expression, light, and pose between target faces and swapped faces can be expressed as:
\begin{equation}
    \begin{split}
        L_{exp} &= MSE(\left( \bm{\psi}_{t} \right),\left( \bm{\psi}_{s}\right)) \\
        L_{light} &= MSE(\left( \mathbf{L}_{t} \right),\left( \mathbf{L}_{s}\right)) \\
        L_{pose} &= Cos(\left( \bm{\theta}_{6D,t} \right),\left( \bm{\theta}_{6D,s} \right))       
    \end{split}
\end{equation}
If there exists such a pair of swapped images \(I_{A}\) and \(I_{B}\), they have the same target image and satisfy:
\begin{equation}
    \left.
        \begin{aligned}
            &L_{A, exp} < L_{B, exp} \\
            &L_{A, light} < L_{B, light} \\
            &L_{A, pose} < L_{B, pose} \\
            &LPIPS(I_{A},I_{T}) < LPIPS(I_{B},I_{T})
        \end{aligned}
        \right\}
        \\
        \quad\Rightarrow\quad
    Q_{A} > Q_{B} 
\end{equation}
With the three losses mentioned above, we generated numerous rank-based labels. To ensure consistency with human perception, we incorporated LPIPS \cite{zhang2018blind} as a further filter for labels. By leveraging all vectors, we created a significant number of ranking labels based on the quality of the face-swap images.

\subsection{Forgery Image Quality Assessment}
After obtaining quality-related classification labels, we propose a Siamese network to learn the ranking in order to obtain more accurate quantitative quality evaluation scores. Unlike the original Siamese network \cite{li2017classification}, the network here learns the classification information, and the loss function is a loss of classification based on the margin. The network has two identical branches that share weights during training. The inputs to the network are image pairs and labels, producing two outputs that are passed to the loss function. Backpropagation calculates the gradient of the loss function with respect to all model vectors, which are updated using stochastic gradient descent. Given an image \(x\) as input, the activation of the last layer produces the output feature representation denoted by \(f(x:\theta)\), where \(\theta\) is the network parameter. \(y\) represents the true value of the image and is used to represent a quality ranking score for objective quality assessment of forged images. In the Siamese network, the output of the last layer is a scalar. Since the goal is to rank images, pairwise ranking loss is used for training. The function of margin-aware ranking loss is as follows:
\begin{equation}
    L\left( x_{1},x_{2};\theta \right) = max(0,f\left( x_{2};\theta \right) - f\left( x_{1};\theta \right) + \varepsilon)
\end{equation}
where \(\varepsilon\) is the boundary setting. For different pairs of forged images \(I_{a}\) and \(I_{b}\), judging only the high and low quality will lead to a particularly large number of sorted forged images, and some of the subtle image quality differences may become errors due to the accuracy of the 3D face reconstruction model. Therefore, \(\varepsilon\) needs to be designed to meet the Siamese network and should be larger than some threshold value for the image quality differences calculated during training.

\section{Assessment for Face Swapping}
\subsection{Datasets}
During the generation of swapped faces, we assembled a dataset comprising $29,336$ images from CelebAMask-HQ \cite{lee2020maskgan}. Each swapped image in the dataset consists of a source image and a target image, both real to maintain consistent quality. The overall dataset includes more than $1$ million images, with more than $140,000$ images representing self-swapped faces derived from five distinct swapping methods. In particular, no real images were included in this subset. To facilitate model training and evaluation, we divided the dataset into training, validation, and test sets, adhering to a 7:2:1 ratio. This division ensures that each phase of model development receives adequate data for effective performance evaluation and learning.

To maximize credibility, we use the DeepFake Game Competition Visual Realism Assessment (DFGC-VRA) database \cite{peng2023dfgc} for the evaluation. This database comprises $1,400$ videos, $20$ pairs of IDs, and $35$ undisclosed face generation methods. Among these, the training set contains $700$ videos, while the test set is divided into three subsets: set C1 (ID disjoint with the training set), set C2 (method disjoint), and set C3 (ID and method disjoint). All videos are annotated with Mean Opinion Scores (MOS). This annotated subset includes $14$ pairs of ID and $25$ face-swapping methods.

In the testing of image quality assessment, to prove the model's generalization in unseen data, all models are conducted without prior exposure to the training set. For each video, we perform uniform sampling of ten frames, predict after obtaining the facial region, and calculate the average of the predicted scores for all sampled frames to derive the final result. This approach ensures a comprehensive assessment of the quality of face-swapping. In the supplementary materials, we also present the results of our model after fine-tuning the training set.

\begin{figure}[tp]
\centering
\includegraphics[width=0.99\linewidth]{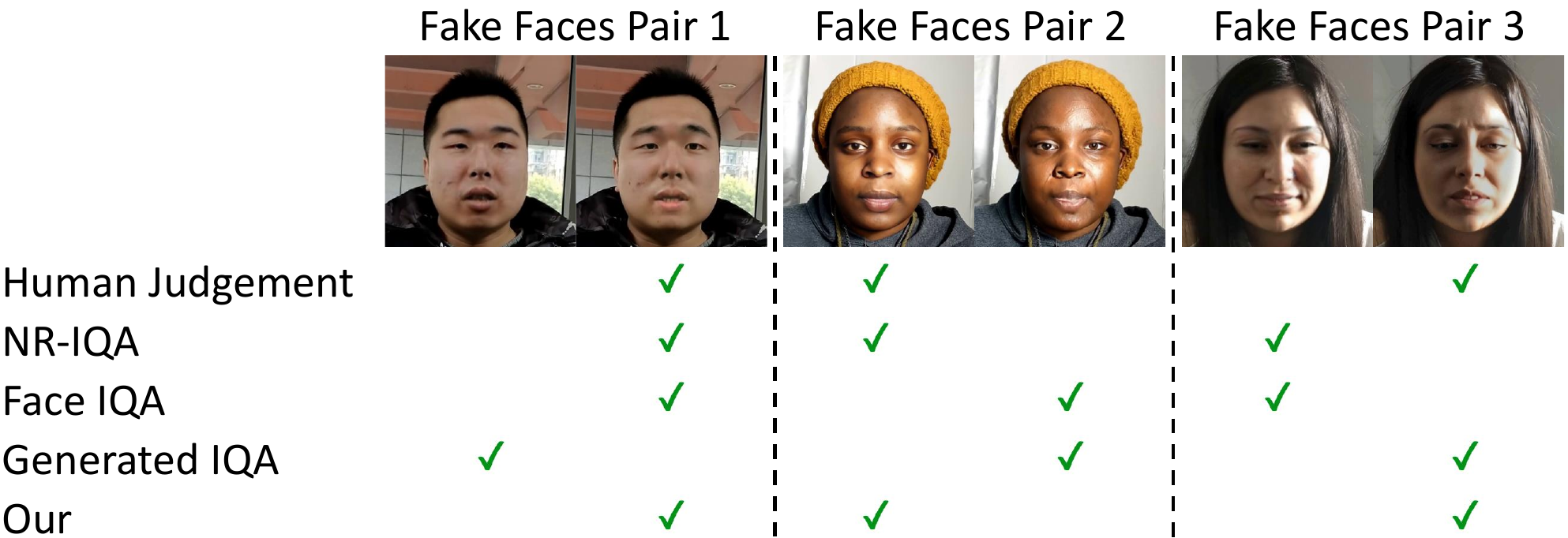}
\caption{The comparison to human judgment, images come from DFGC-VRA \cite{peng2023dfgc}. Relative to other approaches, our method demonstrates greater consistency with human evaluations in assessing the image quality of fake faces. "NR-IQA" refers to MUSIQ \cite{ke2021musiq}, "Face IQA" denotes SDD-FIQA \cite{ou2021sdd}, and "Generate IQA" represents KNN-GIQA \cite{gu2020giqa}.}
\label{fig:tick_comp}
\vspace{-1em}
\end{figure}

\subsection{Experimental protocol}
\textbf{Implementation details.} The implementation of our evaluating model using pytorch, as shown in Table \ref{tab:backbone}, and its performance is influenced by the different backbones, ViT-B/32 \cite{dosovitskiy2020image} provides the best performance. The batch size is set to $32$ and the model is trained for $20$ epochs using two NVIDIA RTX A6000 GPUs. We use Adam with \(\bm{\beta_{1}} = 0\) and \(\bm{\beta_{2}} = 0.99 \). The weight decay and the learning rate are set to 3e-5.

\begin{table}[tb]
    \caption{Consistency (in \%) of NR-IQA and GIQA methods with human judgments. (50\% corresponds to a random guess)}
    \label{tab:human_judgement2}
    \begin{tabularx}{0.99\columnwidth}{@{}l*{6}{>{\centering\arraybackslash}X}@{}}
    \toprule
    \multirow{2}{*}[-0.5ex]{\makecell[cl]{NR-IQA/GIQA}} & \multicolumn{3}{c}{Coarse-grained $\uparrow$} & \multicolumn{3}{c}{Fine-grained $\uparrow$} \\
    \cmidrule(lr){2-7}
     & C1 & C2 & C3 & C1 & C2 & C3 \\
    \midrule
        NIQE \cite{mittal2012making} & 49.63 & 51.14 & 49.07 & 47.45 & 53.72 & 43.20\\
        ILNIQE \cite{zhang2015feature} & 51.18 & 49.34 & 48.90 & 48.22 & 50.97 & 44.08\\
        BRISQUE \cite{mittal2012no}  & 55.97 & 53.50 & 49.34 & 45.18 & 53.59 & 44.32\\
        CNNIQA \cite{kang2014convolutional} & 69.25 & 59.96 & 67.34 & 62.46 & 55.47 & 60.58\\
        NIMA \cite{talebi2018nima}   & 73.90 & 56.05 & 61.76 & 63.17 & 55.48 & 62.13 \\
        PaQ-2-PiQ \cite{ying2020patches} & 71.37 & 61.95 & 66.13 & 63.47 & 56.73 & 64.51 \\
        DBCNN \cite{zhang2018blind}  & 66.71 & 55.27 & 65.67 & 59.68 & 55.11 & 57.99 \\
        HyperIQA \cite{su2020blindly} & 71.67 & 58.37 & 65.04 & 70.37 & 58.54 & 64.28 \\
        MUSIQ \cite{ke2021musiq} & 70.38 & 61.21 & 66.01 & 60.43 & 57.56 & 65.23\\
        Tres \cite{golestaneh2022no} & 71.16 & 55.86 & 63.49 & 67.33 & 55.10 & 61.95 \\
        MANIQA \cite{yang2022maniqa} & 71.23 & 61.87 & 67.54 & 61.22 & 60.42 & 66.84 \\
        CLIPIQA \cite{wang2022exploring} & 61.07 & 51.36 & 69.84 & 57.12 & 51.81 & 67.04 \\
        LIQE \cite{zhang2023blind} & 78.28 & 60.14 & \textbf{70.43} & 67.95 & 55.30 & \textbf{67.92} \\
        \hline
        GMM-GIQA \cite{gu2020giqa} &50.28 & 50.21 & 50.15 & 49.76 & 51.06 & 49.37 \\
        KNN-GIQA \cite{gu2020giqa} &62.59 & 60.78 & 61.02 & 54.75 & 53.25 & 55.67 \\
        \hline
        \textbf{Ours} & \textbf{85.75} & \textbf{67.15} & 69.50 & \textbf{73.78} & \textbf{63.38} & 67.47 \\
    \bottomrule
    \end{tabularx}
\end{table}

\begin{table}[tb]
    \caption{Consistency (in \%) of FIQA methods with human judgments. (50\% corresponds to a random guess)}
    \label{tab:human_judgement1}
    \begin{tabularx}{0.99\columnwidth}{@{}l*{6}{>{\centering\arraybackslash}X}@{}}
    \toprule
    \multirow{2}{*}[-0.5ex]{\makecell[cl]{FIQA}} & \multicolumn{3}{c}{Coarse-grained$\uparrow$} & \multicolumn{3}{c}{Fine-grained $\uparrow$} \\
    \cmidrule(lr){2-7}
     & C1 & C2 & C3 &  C1 & C2 & C3\\
    \midrule
    FaceQnet \cite{hernandez2019faceqnet} & 50.75 & 51.66 & 48.97 & 50.18 & 51.22 & 50.06 \\
    SER-FIQ \cite{terhorst2020ser} & 51.95 & 53.23 & 50.96 & 47.65 & 50.05 & 48.97 \\
    MagFace \cite{meng2021magface} & 56.26 & 55.04 & 53.43 & 49.25 & 50.03 & 49.99 \\
    SDD-FIQA \cite{ou2021sdd} & 58.50 & 56.45 & 53.98 & 54.24 & 54.13 & 53.11  \\
    CR-FIQA \cite{boutros2023cr} & 47.18 & 50.01 & 48.78 & 48.63 & 51.46 & 52.01 \\
    IFQA \cite{jo2023ifqa} & 49.49 & 50.34 & 48.90 & 49.83 & 50.02 & 50.32 \\
    \hline      
    \textbf{Ours} & \textbf{85.75} & \textbf{67.15} & \textbf{69.50} & \textbf{73.78} & \textbf{63.38} & \textbf{67.47} \\
    \bottomrule
    \end{tabularx}
    \vspace{-9pt}
\end{table}

\textbf{Human evaluation.}
To evaluate the consistency of each metric with human judgment, we performed binary classification experiments. Each test sample consisted of a reference image and a video randomly sampled from the DFGC-VRA dataset \cite{peng2023dfgc}. Of the 700 videos with independent human ratings, we used each video's evaluation score for comparison with other videos to ensure experimental diversity. Controversial samples were removed, leaving 672 samples for metric evaluation. To accurately assess the quality evaluation model's performance, we divided the samples into coarse-grained groups (paired videos with human evaluation differences of 1 or less) and fine-grained groups (paired videos with human evaluation differences of 0.1 or less). We compared our approach to a baseline of 16 metrics, including nine NR-IQA methods, six FIQA methods, and two GIQA methods.

\subsection{Ablation Studies}
\textbf{Ablation of attributes.} We examine each component of our framework in Table \ref{tab:abalation1}, where each component is cumulatively added to the generation conditions of the classification labels. We find that when considering only two attributes, pose and expression, our metric achieves a base performance sufficient to give more than satisfactory predictions for the test sample. Lighting and perceptual consistency are non-relevant attributes for faces but are still crucial for evaluating the image quality of swapped faces.
\begin{table}[tb]
    \caption{Performance of different network backbones.}
      \label{tab:backbone}
        \begin{tabular}{lcc}
        \toprule
        Backbone & Coarse-grained $\uparrow$ & Fine-grained $\uparrow$ \\
        \midrule
        MobileNetV3-S        & 70.88        & 63.70        \\
        MobileNetV3-L        & 73.50        & 66.54        \\
        ResNet50       & 73.74        & 67.11        \\
        ResNet152          & 74.25        & 67.88      \\
        Xception           & 73.53        & 67.02      \\
        EfficientNet-B0       & 75.79        & 68.55      \\   
        \textbf{ViT-B/32} & \textbf{75.85} & \textbf{68.63} \\
    \bottomrule
  \end{tabular}
\end{table}
\begin{table}[tb]
    \caption{Effectiveness of various attributes.}
      \label{tab:abalation1}
        \begin{tabular}{lcc}
        \toprule
        Method & Coarse-grained $\uparrow$ & Fine-grained $\uparrow$\\
        \midrule
        Only Expression        & 54.63        & 51.91        \\
        Only Pose        & 52.24        & 50.60        \\
        Only lighting        & 51.08        & 50.44        \\
        \hline
        Pose\&Expression        & 70.26        & 60.48        \\
        +lighting        & 71.38        & 64.82        \\
        +LPIPS  & 73.13 & 66.15        \\
        \textbf{+LPIPS + Light}       & \textbf{75.85}        & \textbf{68.63}       \\
    \bottomrule
  \end{tabular}
\end{table}

\textbf{ID similarity in Rank-based Quality Assessment.}
Table \ref{tab:abalation2} compares the impact of adding identity similarity to the current attribute-based quality assessment model. This is done by adding new labels to the ranking labels that satisfy the following conditions:
\begin{equation}
    L_{id}(I_{a}, I_{r}) < L_{id}(I_{b}, I_{r}) \rightarrow Q_{a} > Q_{b}
\end{equation}
where:
\begin{equation}
    L_{id} = 1 - \frac{f(I)f(I_r)}{\left\| f(I) \right\|_2 \cdot \left\| f(I_r) \right\|_2} 
\end{equation}
\(f\) is a face classification model \cite{kim2022adaface} we used, it provides some inspirational comments for the quality of face images.

\begin{table}[tb]
    \caption{Effectiveness of adding identity similarity.}
      \label{tab:abalation2}
        \begin{tabular}{lcc}
        \toprule
        Method & Coarse-grained $\uparrow$ & Fine-grained $\uparrow$\\
        \midrule
        Pose\&Expression     & 70.26        & 60.48        \\
        +ID Similarity       & 70.31        & 60.67        \\
        +Light+ID Similarity & 70.61        & 61.34        \\
        \textbf{+LPIPS +ID Similarity} & \textbf{71.62}      & \textbf{64.91} \\
    \bottomrule
  \end{tabular}
  \vspace{-5pt}
\end{table}

\textbf{Dataset Magnitude.}
In our experiments, we found that the performance of the evaluation model is positively related to the size of the training set. As the number of images gradually reduces from the original $50\%$ to $10\%$, our metric will give random results for all test data. On the other hand, when the number of trained images reaches around 800K, more data will no longer significantly improve performance. This experiment goes some way to explaining the large-scale face-swapping dataset constructed in this paper.
\begin{table}[tb]
    \caption{Effectiveness of the size of the training set.}
      \label{tab:abalation3}
        \begin{tabular}{lcc}
        \toprule
        Scales of Training Set & Coarse-grained $\uparrow$ & Fine-grained $\uparrow$\\
        \midrule
        Random                                  & 50.00        & 50.00        \\
        10\%(\textasciitilde100K images)        & 51.96        & 49.87        \\
        30\%(\textasciitilde250K images)        & 62.52        & 58.71        \\
        50\%(\textasciitilde400K images)        & 64.98        & 63.81        \\
        70\%(\textasciitilde550K images)        & 73.88        & 66.96        \\
        90\%(\textasciitilde700K images)        & 75.53        & 68.48      \\    
        \textbf{100\%(\textasciitilde800K images)}        & \textbf{75.85}        & \textbf{68.63}       \\   
    \bottomrule
  \end{tabular}
\end{table}

\begin{figure*}[htb]
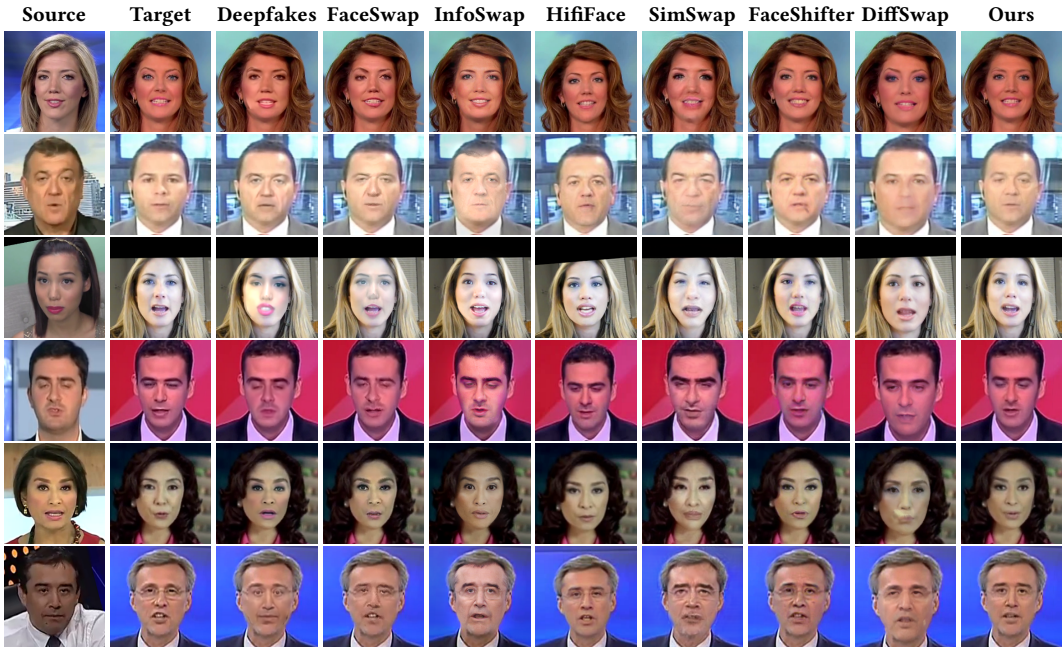

\centering
\foreach \row in {01,02,03,04,05,06}{
    \ifnum\row=01
        \foreach \col/\name in {0src/Source, 0tgt/Target, df/Deepfakes, fs/FaceSwap, is/InfoSwap, hf/HifiFace, sw/SimSwap, fsh/FaceShifter, diff/DiffSwap, ours/Ours}{
            \begin{subfigure}{0.075\textwidth}
                \centering
                \caption*{\textbf{\name}} 
                \includegraphics[width=\linewidth]{image_qualitative/candi_\row/\col.png}
            \end{subfigure}
        }
    \else
        \foreach \col in {0src,0tgt,df,fs,is,hf,sw,fsh,diff,ours}{
            \begin{subfigure}{0.075\textwidth}
                \centering
                \includegraphics[width=\linewidth]{image_qualitative/candi_\row/\col.png}
            \end{subfigure}
        }
    \fi
    \par
}
\caption{Qualitative comparison on FaceForensics++ \cite{rossler2019faceforensics++}. Our model is capable of achieving precise face swapping while preserving target attributes such as expressions and poses. More results can be found in the supplementary materials.}
\label{qualitative comp}
\end{figure*}

\section{Improving Face-Swapping with Quality}
\subsection{Training Loss}
To better validate the effectiveness of our quality metrics $M$, we refined a face swapping model that is more aligned with human perception, based on a known state-of-the-art architecture. AEI-Net \cite{li2020advancing}. This task requires generating a satisfactory result \(Y_{s,t}\) by manipulating a source face \(X_{s}\) and a target face \(X_{t}\). To meet the requirements, we incorporate a quality assessment error in the form of L1Loss into the generated faces, with the aim of generating a fake face \(\hat{Y}_{s,t}\) to maintain consistency with the target face \(X_{t}\) in terms of image quality assessment. Specifically, this loss function is expressed as follows:
\begin{equation}
    L_{quality} = |M({X}_{t})-M(\hat{Y}_{s,t})|
\end{equation}
\noindent Furthermore, we employ $L_{adv}$ as the adversarial loss to render $\hat{Y}_{s,t}$ realistic. It is implemented as a multiscale discriminator \cite{park2019semantic} on the faces generated from downsampling. Moreover, an identity preservation loss is utilized to maintain the identity of the source.
\begin{equation}
    L_{id} = 1 - cos(\bm{z}_{id}(Y_{s, t}), \bm{z}_{id}(X_s))
\end{equation}
where $\bm{z}_{id}$ is the identity vector extracted from  face recognition model \cite{deng2019accurate}. We also employ $L_{rec}$ a reconstruction loss as pixel level $L$-2 distances between the target image $X_t$ and $\hat{Y}_{s,t}$:
\begin{equation}
\label{eqn:L_rec}
L_{rec} = 
\begin{cases}
\frac{1}{2} \left\|\hat{Y}_{s,t} - X_t\right\|_2^2 & \textit{if}~X_t = X_s \\
0 & \textit{otherwise}
\end{cases}
\end{equation}
\noindent A finer attribute preservation loss $L_{attr}$ is defined in the original training loss. To avoid unnecessary interference, we refrained from utilizing this loss function. Related ablation studies are available in the supplementary materials.

The improved face swapping is ﬁnally trained with a weighted sum of above losses as:
\begin{equation}
\label{eqn:L_all}
    L_{imporved} = L_{adv} + \lambda_{1} L_{id}  + \lambda_{2} L_{rec} + \lambda_{3} L_{quality}
\end{equation}
with $\lambda_{1}=20$, $\lambda_{2}=7$, $\lambda_{3}=0.25$. All other parameters of the experiment were kept consistent with the original architecture.

\begin{table}[tb]
    \caption{Quantitative comparison on FaceForensics++ \cite{rossler2019faceforensics++}. The best result is highlighted in boldface, and the second best result is highlighted in underline.}
      \label{tab:quanti comp}
        \begin{tabular}{lccc}
        \toprule
        Method & ID Retrieval $\uparrow$ & Pose $\downarrow$ & Expression $\downarrow$\\
        \midrule
        DeepFakes \cite{Deepfakes}   & 94.09 & 3.51 & 2.98 \\
        FaceSwap \cite{FaceSwap}   & 82.60 & 2.96 & 2.61 \\
        SimSwap \cite{chen2020simswap}     & 95.90 & 3.21 & 2.88 \\
        FaceShifter \cite{li2020advancing} & 95.38 & \underline{2.94} & \underline{2.48} \\
        HifiFace \cite{wang2021hififace}   & \underline{97.79} & 3.42 & 2.57 \\
        InfoSwap \cite{gao2021information}   & 96.89 & 3.06 & 2.89 \\
        MegaFS \cite{zhu2021one} & 92.45 & 3.21 & 2.63 \\
        DiffSwap \cite{zhao2023diffswap}   & \textbf{98.54} & 2.95 & 2.68 \\
        \midrule
        Ours & 95.25 & \textbf{2.57} & \textbf{2.32} \\
    \bottomrule
  \end{tabular}
\end{table}

\subsection{Qualitative and Quantitative Comparisons}
\textbf{Qualitative Comparison.} We compare our method with Deepfakes \cite{Deepfakes}, FaceSwap \cite{FaceSwap}, InfoSwap \cite{gao2021information}, HifiFace \cite{wang2021hififace}, SimSwap \cite{chen2020simswap}, MegaFS \cite{zhu2021one} and DiffSwap \cite{zhao2023diffswap} in FaceForensics++. As illustrated in Figure \ref{qualitative comp}, our approach preserves pose, expression, and gaze more effectively, surpassing previous methods.

\textbf{Quantitative Comparison.}
Following the experimental setup in FaceShifter \cite{li2020advancing}, we uniformly sample 10 frames of each video and obtain faces using MTCNN \cite{zhang2016joint}. We applied a different advanced face recognition model AdaFace \cite{kim2022adaface} to extract the embedding of the identity and retrieve the closest face using cosine similarity. A pose estimator \cite{ruiz2018fine} and a 3D facial model \cite{deng2019accurate} are used to extract pose and expression vectors for pose and expression evaluation. Compared to the baseline model \cite{li2020advancing}, the pose error and the expression error of our method improved by 0.37 ($\textbf{12.5\%}$ improvement) and 0.16 ($\textbf{6.4\%}$ improvement), respectively, with a small decrease of $0.13\%$ in ID Retrieval. The small decrease in ID retrieval may be due to the fact that existing face-swapping methods give excessive attention to the target's facial identity while ignoring other attributes that are also important for subjective human perception, while the introduction of the proposed assessment metrics as loss constraints rectifies this point well, leading to better visual results of face-swapping, as shown in Figure \ref{qualitative comp}.

\vspace{-10pt}

\section{Conclusion}
In conclusion, we introduce a rank-based no-reference quality assessment metric tailored for face-swapping applications. By tackling the prevalent challenges of anonymity and unrealistic distortions head on, we have developed a comprehensive, large-scale dataset and a novel methodology for ranking image quality based on a multitude of facial attributes. Our approach uses a Siamese network architecture to facilitate qualitative interpretable comparisons, setting a new benchmark to assess the quality of swapped faces. Our metric excels at providing both broad and detailed evaluations, outperforming existing no-reference image quality assessment metrics and the most advanced face image quality assessment metrics. Extensive experimental validation underscores the superiority of our method, establishing it as a highly effective tool for evaluating face-swapping images in practical scenarios. By improving the accuracy and depth of face replacement image evaluations, our metric significantly increases the credibility of face-swapping technologies.
\bibliographystyle{ACM-Reference-Format}
\bibliography{main}

\end{document}


\title{Rank-based No-reference Quality Assessment for Face Swapping Supplementary Materials}

\maketitle


\begin{figure}[htb]
\centering

\begin{subfigure}{0.15\textwidth}
\centering
\includegraphics[width=\linewidth]{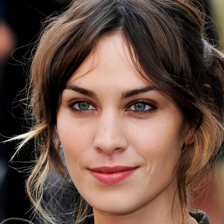}
\caption{Real}
\end{subfigure}
\hfill
\begin{subfigure}{0.15\textwidth}
\centering
\includegraphics[width=\linewidth]{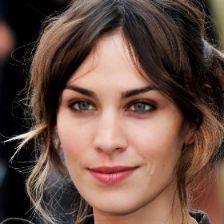}
\caption{FaceShifter}
\end{subfigure}
\hfill
\begin{subfigure}{0.15\textwidth}
\centering
\includegraphics[width=\linewidth]{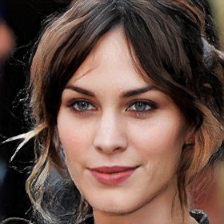}
\caption{SimSwap}
\end{subfigure}

\medskip

\begin{subfigure}{0.15\textwidth}
\centering
\includegraphics[width=\linewidth]{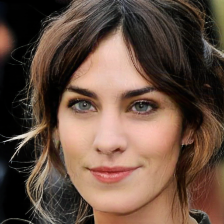}
\caption{InfoSwap}
\end{subfigure}
\hfill
\begin{subfigure}{0.15\textwidth}
\centering
\includegraphics[width=\linewidth]{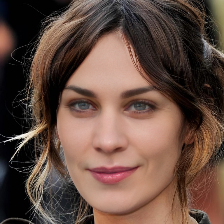}
\caption{MegaFS}
\end{subfigure}
\hfill
\begin{subfigure}{0.15\textwidth}
\centering
\includegraphics[width=\linewidth]{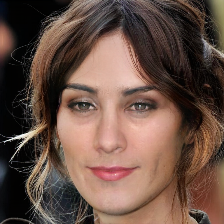}
\caption{HiRes}
\end{subfigure}

\caption{Self-swapping faces in our dataset, with the same source and target, show a lower distortion and a higher quality. We use these high-quality images as real face substitutes, ensuring rank-based label reliability for fake faces. After incorporating self-swapped images into the dataset, we obtain more labels from high-quality swapping faces.}
\vspace{-0.6cm}
\label{fig:self_swapping}
\end{figure}

\section{Details of the rank-based dataset.}
\subsection{Images}
In order to avoid much manual labeling work while effectively representing human subjective perception, we suggest using the inconsistencies between different artificially generated faces with the same target to create reliable hierarchical pseudo-labels. The data we use in our study comes from the CelebAMask-HQ dataset, which includes 29,336 images with identifiable facial features and measurable attributes, each image having a resolution of $1024\times1024$. Additionally, we employ five different methods to create synthetic faces. Please refer to Table ~\ref{tab:data_num} for specific data distributions.

The dataset we created is the first work of no-reference quality assessment for face-swapping, characterized by its large scale, rich rank-based annotations, and high-resolution faces. Compared to other publicly available face forgery datasets, although our dataset is not the largest in scale, it possesses the highest number of high-resolution forged images (with resolutions exceeding $512\times512$), as detailed in Table \ref{tab:dataset_compare}.

\begin{table}[tb]
\caption{Images count by method in our dataset.}
\begin{tabular}{lccc}
\toprule
Method & Resolution & Cross-Swap & Self-Swap\\
\midrule
FaceShifter & \(256 \times 256\) & 182,438 & 29,145 \\
SimSwap & \(256 \times 256\) & 283,719 & 29,366 \\
InfoSwap & \(512 \times 512\) & 287,180 & 29,366 \\
MegaFS & \(1024 \times 1024\) & 238,775 & 29,366 \\
HiRes & \(1024 \times 1024\) & 99,961 & 29,366 \\
\midrule
All & & 1,092,073 & 146,609 \\
\bottomrule
\end{tabular}
\label{tab:data_num}
\end{table}

The purpose of self-swapping images, as illustrated in Figure \ref{fig:self_swapping}, is to ensure that the fake faces closest to the real face receive the highest label in rank-based datasets, rather than the real image itself. Intervention of real faces will allow the quality assessment model to learn about image distortions from real scenes.

\begin{table}[tb]
\caption{Comparison with other publicly available image-level datasets on face swapping and deepfake.}
\centering
\begin{tabular}{lccc}
\toprule
Dataset   & Real& Fake& Fake-HQ     \\
\midrule
UADFV     & 241 & 252 & - \\
SwapMe \& FaceSwap & 4600& 2010& - \\
DFFD      & 58,703        & 240,336       & - \\
ForgeryNet  & \textbf{1,438,201} & \textbf{1,457,861} & 295,526     \\
\textbf{Ours}         & 29,366        & 1,238,682     & \textbf{714,014} \\
\bottomrule
\end{tabular}
\label{tab:dataset_compare}
\end{table}

\begin{table}[h]
\caption{Number of labels and images for different splits.}
\centering
\begin{tabular}{lcc}
\toprule
Split   & Rank-based Labels & Rank-based Images \\
\midrule
Training & 2,764,196 & 1,222,779 \\
Validation & 789,770 & 844,691 \\
Testing & 394,886 & 558,169 \\
\bottomrule
\end{tabular}
\label{tab:label_num}
\end{table}

\begin{figure*}[htp]
\begin{flushleft}
\noindent
\begin{minipage}{0.12\textwidth}
    \caption*{\Large FaceShifter}
    \vspace{6pt}
    \caption*{source ID}
\end{minipage}%
\begin{minipage}{0.08\textwidth}
    \includegraphics[width=\linewidth]{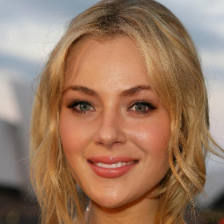}
    \vspace{-20pt}
    \caption*{Self}
\end{minipage}%
\begin{minipage}{0.08\textwidth}
    \includegraphics[width=\linewidth]{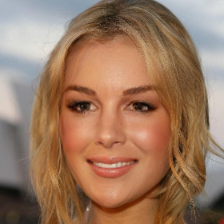}
    \vspace{-20pt}
    \caption*{02752}
\end{minipage}%
\begin{minipage}{0.08\textwidth}
    \includegraphics[width=\linewidth]{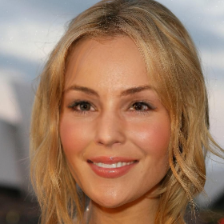}
    \vspace{-20pt}
    \caption*{03157}
\end{minipage}%
\begin{minipage}{0.08\textwidth}
    \includegraphics[width=\linewidth]{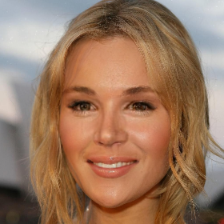}
    \vspace{-20pt}
    \caption*{08171}
\end{minipage}%
\begin{minipage}{0.08\textwidth}
    \includegraphics[width=\linewidth]{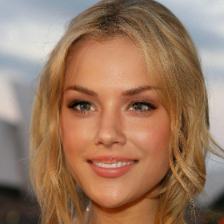}
    \vspace{-20pt}
    \caption*{20512}
\end{minipage}%
\begin{minipage}{0.08\textwidth}
    \includegraphics[width=\linewidth]{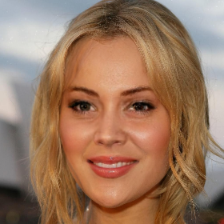}
    \vspace{-20pt}
    \caption*{24199}
\end{minipage}%
\begin{minipage}{0.08\textwidth}
    \includegraphics[width=\linewidth]{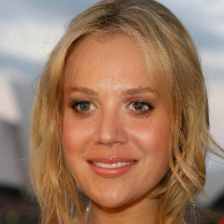}
    \vspace{-20pt}
    \caption*{24666}
\end{minipage}%
\begin{minipage}{0.08\textwidth}
    \includegraphics[width=\linewidth]{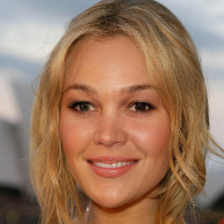}
    \vspace{-20pt}
    \caption*{26444}
\end{minipage}%
\\
\begin{minipage}{0.12\textwidth}
    \caption*{\Large SimSwap}
    \vspace{6pt}
    \caption*{source ID}
\end{minipage}%
\begin{minipage}{0.08\textwidth}
    \includegraphics[width=\linewidth]{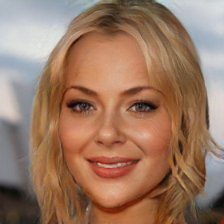}
    \vspace{-20pt}
    \caption*{Self}
\end{minipage}%
\begin{minipage}{0.08\textwidth}
    \includegraphics[width=\linewidth]{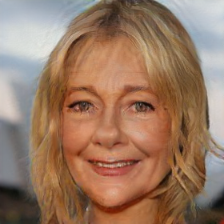}
    \vspace{-20pt}
    \caption*{03290}
\end{minipage}%
\begin{minipage}{0.08\textwidth}
    \includegraphics[width=\linewidth]{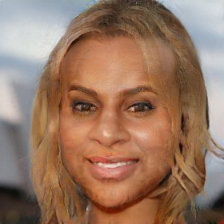}
    \vspace{-20pt}
    \caption*{10456}
\end{minipage}%
\begin{minipage}{0.08\textwidth}
    \includegraphics[width=\linewidth]{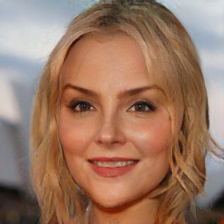}
    \vspace{-20pt}
    \caption*{12205}
\end{minipage}%
\begin{minipage}{0.08\textwidth}
    \includegraphics[width=\linewidth]{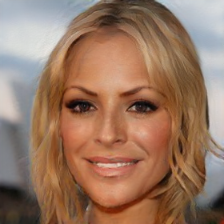}
    \vspace{-20pt}
    \caption*{16409}
\end{minipage}%
\begin{minipage}{0.08\textwidth}
    \includegraphics[width=\linewidth]{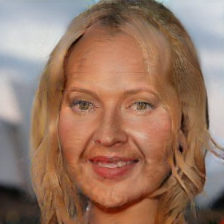}
    \vspace{-20pt}
    \caption*{23050}
\end{minipage}%
\begin{minipage}{0.08\textwidth}
    \includegraphics[width=\linewidth]{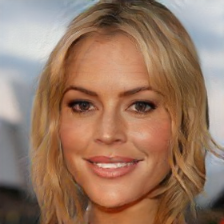}
    \vspace{-20pt}
    \caption*{24303}
\end{minipage}%
\begin{minipage}{0.08\textwidth}
    \includegraphics[width=\linewidth]{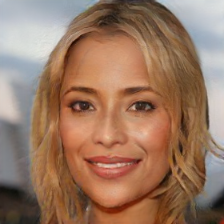}
    \vspace{-20pt}
    \caption*{28924}
\end{minipage}%
\\
\begin{minipage}{0.12\textwidth}
    \caption*{\Large InfoSwap}
    \vspace{6pt}
    \caption*{source ID}
\end{minipage}%
\begin{minipage}{0.08\textwidth}
    \includegraphics[width=\linewidth]{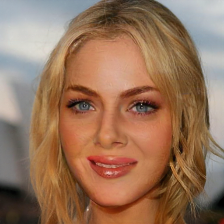}
    \vspace{-20pt}
    \caption*{Self}
\end{minipage}%
\begin{minipage}{0.08\textwidth}
    \includegraphics[width=\linewidth]{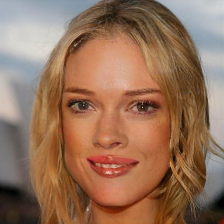}
    \vspace{-20pt}
    \caption*{06400}
\end{minipage}%
\begin{minipage}{0.08\textwidth}
    \includegraphics[width=\linewidth]{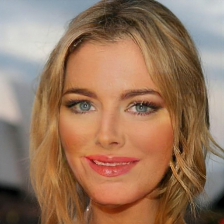}
    \vspace{-20pt}
    \caption*{06535}
\end{minipage}%
\begin{minipage}{0.08\textwidth}
    \includegraphics[width=\linewidth]{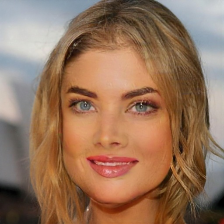}
    \vspace{-20pt}
    \caption*{09669}
\end{minipage}%
\begin{minipage}{0.08\textwidth}
    \includegraphics[width=\linewidth]{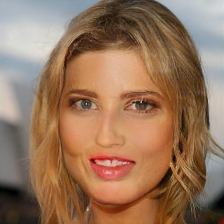}
    \vspace{-20pt}
    \caption*{10165}
\end{minipage}%
\begin{minipage}{0.08\textwidth}
    \includegraphics[width=\linewidth]{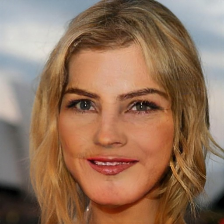}
    \vspace{-20pt}
    \caption*{11595}
\end{minipage}%
\begin{minipage}{0.08\textwidth}
    \includegraphics[width=\linewidth]{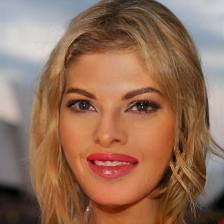}
    \vspace{-20pt}
    \caption*{12527}
\end{minipage}%
\begin{minipage}{0.08\textwidth}
    \includegraphics[width=\linewidth]{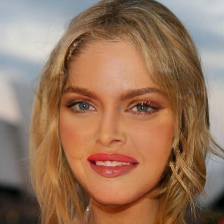}
    \vspace{-20pt}
    \caption*{13998}
\end{minipage}%
\begin{minipage}{0.08\textwidth}
    \includegraphics[width=\linewidth]{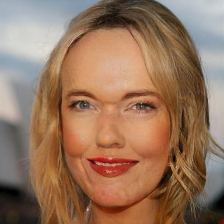}
    \vspace{-20pt}
    \caption*{14774}
\end{minipage}%
\begin{minipage}{0.08\textwidth}
    \includegraphics[width=\linewidth]{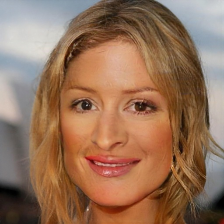}
    \vspace{-20pt}
    \caption*{20504}
\end{minipage}%
\begin{minipage}{0.08\textwidth}
    \includegraphics[width=\linewidth]{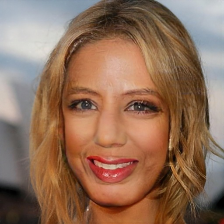}
    \vspace{-20pt}
    \caption*{21193}
\end{minipage}%
\\
\begin{minipage}{0.12\textwidth}
    \caption*{\Large MegaFS}
    \vspace{6pt}
    \caption*{source ID}
\end{minipage}%
\begin{minipage}{0.08\textwidth}
    \includegraphics[width=\linewidth]{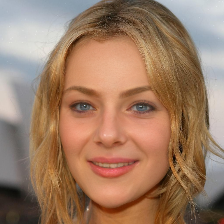}
    \vspace{-20pt}
    \caption*{Self}
\end{minipage}%
\begin{minipage}{0.08\textwidth}
    \includegraphics[width=\linewidth]{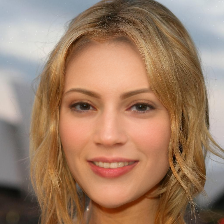}
    \vspace{-20pt}
    \caption*{06972}
\end{minipage}%
\begin{minipage}{0.08\textwidth}
    \includegraphics[width=\linewidth]{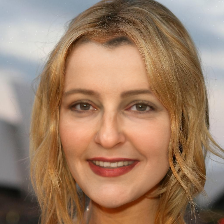}
    \vspace{-20pt}
    \caption*{10274}
\end{minipage}%
\begin{minipage}{0.08\textwidth}
    \includegraphics[width=\linewidth]{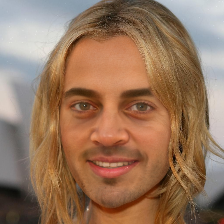}
    \vspace{-20pt}
    \caption*{12295}
\end{minipage}%
\begin{minipage}{0.08\textwidth}
    \includegraphics[width=\linewidth]{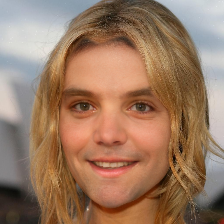}
    \vspace{-20pt}
    \caption*{14483}
\end{minipage}%
\begin{minipage}{0.08\textwidth}
    \includegraphics[width=\linewidth]{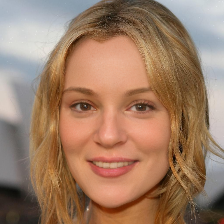}
    \vspace{-20pt}
    \caption*{16270}
\end{minipage}%
\begin{minipage}{0.08\textwidth}
    \includegraphics[width=\linewidth]{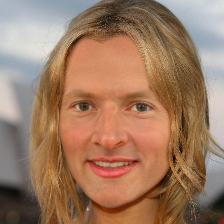}
    \vspace{-20pt}
    \caption*{17194}
\end{minipage}%
\begin{minipage}{0.08\textwidth}
    \includegraphics[width=\linewidth]{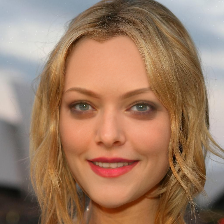}
    \vspace{-20pt}
    \caption*{22683}
\end{minipage}%
\begin{minipage}{0.08\textwidth}
    \includegraphics[width=\linewidth]{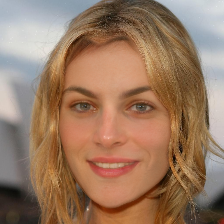}
    \vspace{-20pt}
    \caption*{24039}
\end{minipage}%
\begin{minipage}{0.08\textwidth}
    \includegraphics[width=\linewidth]{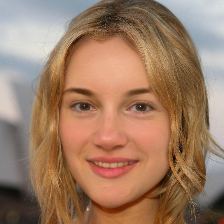}
    \vspace{-20pt}
    \caption*{25415}
\end{minipage}%
\begin{minipage}{0.08\textwidth}
    \includegraphics[width=\linewidth]{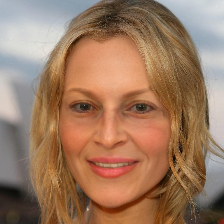}
    \vspace{-20pt}
    \caption*{25803}
\end{minipage}%
\\
\begin{minipage}{0.12\textwidth}
    \caption*{\Large HiRes}
    \vspace{6pt}
    \caption*{source ID}
\end{minipage}%
\begin{minipage}{0.08\textwidth}
    \includegraphics[width=\linewidth]{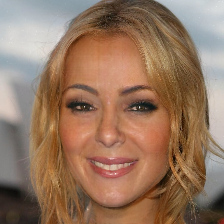}
    \vspace{-20pt}
    \caption*{Self}
\end{minipage}%
\begin{minipage}{0.08\textwidth}
    \includegraphics[width=\linewidth]{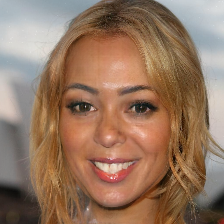}
    \vspace{-20pt}
    \caption*{04358}
\end{minipage}%
\begin{minipage}{0.08\textwidth}
    \includegraphics[width=\linewidth]{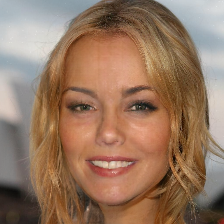}
    \vspace{-20pt}
    \caption*{04631}
\end{minipage}%
\begin{minipage}{0.08\textwidth}
    \includegraphics[width=\linewidth]{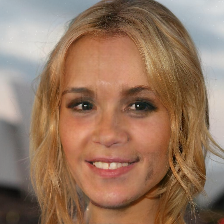}
    \vspace{-20pt}
    \caption*{05536}
\end{minipage}%
\begin{minipage}{0.08\textwidth}
    \includegraphics[width=\linewidth]{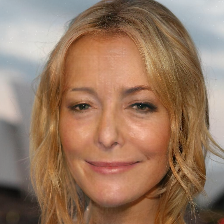}
    \vspace{-20pt}
    \caption*{10407}
\end{minipage}%
\begin{minipage}{0.08\textwidth}
    \includegraphics[width=\linewidth]{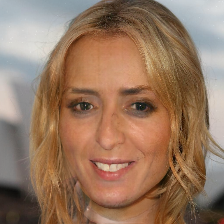}
    \vspace{-20pt}
    \caption*{13714}
\end{minipage}%
\begin{minipage}{0.08\textwidth}
    \includegraphics[width=\linewidth]{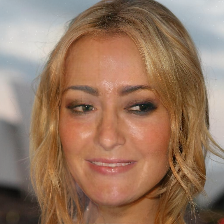}
    \vspace{-20pt}
    \caption*{14679}
\end{minipage}%
\begin{minipage}{0.08\textwidth}
    \includegraphics[width=\linewidth]{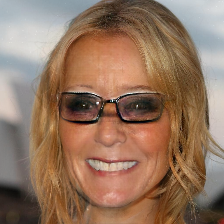}
    \vspace{-20pt}
    \caption*{18062}
\end{minipage}%
\begin{minipage}{0.08\textwidth}
    \includegraphics[width=\linewidth]{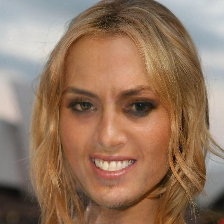}
    \vspace{-20pt}
    \caption*{18906}
\end{minipage}%
\begin{minipage}{0.08\textwidth}
    \includegraphics[width=\linewidth]{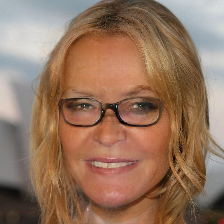}
    \vspace{-20pt}
    \caption*{26721}
\end{minipage}%
\caption{Swapped images with target ID $\textbf{00122}$. These faces are generated from different source faces (the source ID is below the image). Few swapped face images cannot be extracted from all facial attributes, so the numbers of images from each face-swapping method are different.} 
\label{fig:template_images}
\end{flushleft}
\vspace{-10pt}
\end{figure*}

To introduce our rank-based dataset more intuitively, we randomly select a set of fake faces with the same target face, and the ID of the source is labeled below the image, as shown in Figure \ref{fig:template_images}.

\begin{figure*}[h]
\centering
\includegraphics[width=0.83\linewidth]{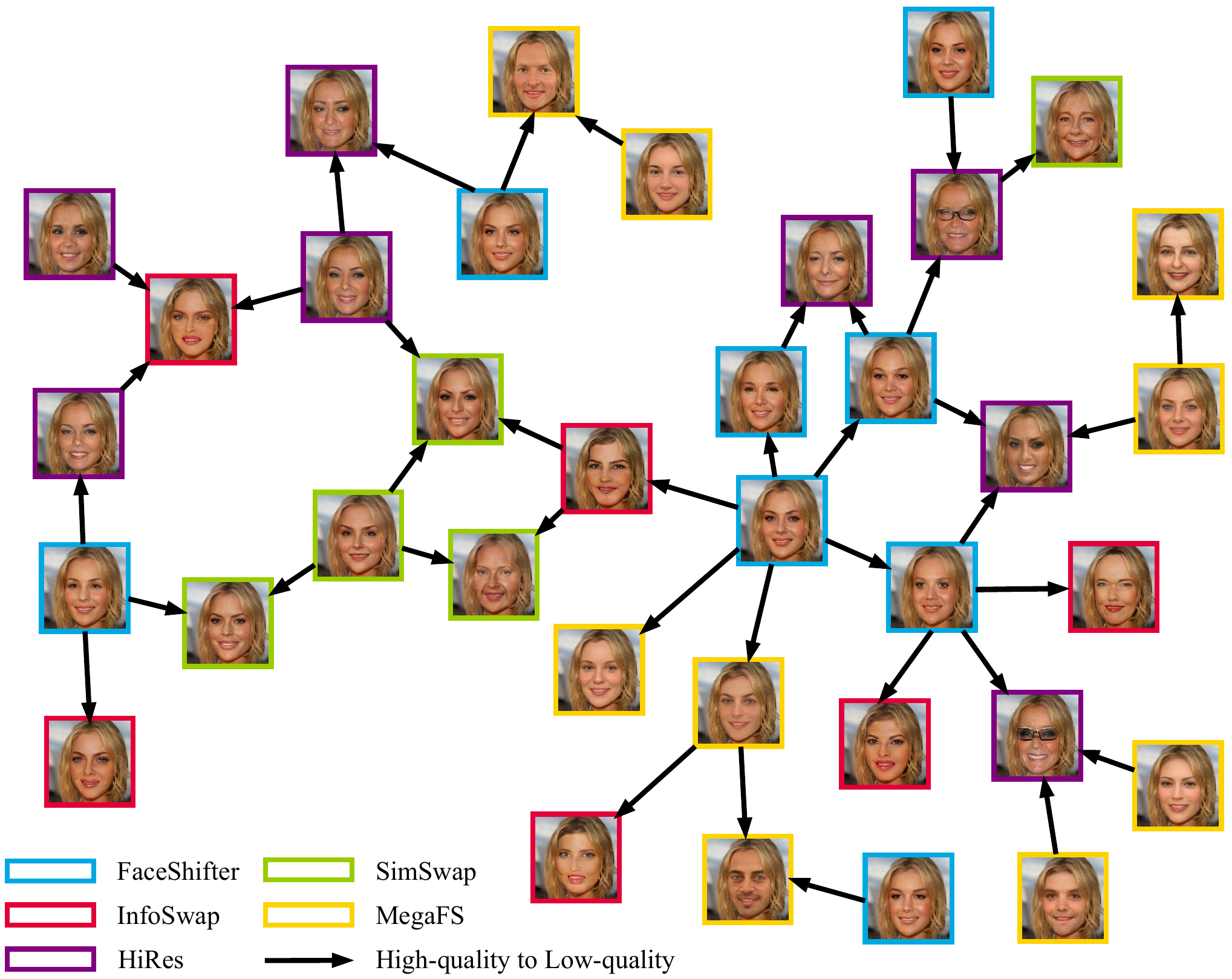}
\caption{A quality rank-based directed graph with five face swapping methods in our dataset. All fake faces are sampled from Figure \ref{fig:template_images}, swapped by target ID $\textbf{00122}$. We represent the image generation methods using different colors and connect high-quality faces to low-quality faces with arrows. All labels in the dataset can be represented as such directed graphs.}
\label{fig:directed_graph}.
\vspace{-10pt}
\end{figure*}

The dataset consists of both low- and high-resolution images, and the data sources are well balanced, with no clear preference. Compared to publicly available counterfeit image datasets, our dataset has the largest number of high-resolution counterfeit images, which is crucial for forging face quality assessment.

\subsection{Labels}
Due to the special attributes of the data labels, all rank-based labels are divided into three parts by number, and the number of images in different splits does not correspond to the number of labels, as shown in Table ~\ref{tab:label_num}. The total number of images in the training set is close to all the data, for example: there are labels $A>B$ and $C>D$ in the training set, and the model cannot determine labels such as $B>C$ and $A>D$ in the validation set.

To understand the ranking relationships in the dataset, we illustrate a directed graph with quality-ranked labels and a real target face in Figure \ref{fig:directed_graph}. The direction between nodes indicates a transition from high to low quality, with images sourced from Figure \ref{fig:template_images}. Upon analysis, most images show a quality distribution centered around the mean, with few exhibiting extreme qualities. This pattern aligns with human judgment, suggesting that our method effectively mirrors general image quality perception.

\begin{figure*}[htb]
\centering
\includegraphics[width=0.8\linewidth]{rank_label_ffpp.pdf}
\caption{Compared to other IQA methods, (a) means comparison with FIQA, (b) means comparison with NR-IQA. Our quality evaluation metric can well rank the quality of FaceForensics++'s fake faces in line with human perception. This shows that our model can be well applied to unseen faces.}
\label{fig:rank_label_ffpp}
\end{figure*}


\section{Predicting ranked labels in FaceForensics++.}
As illustrated in Figure \ref{fig:rank_label_ffpp}, our metric yields rankings aligned with human perception. Initially, fake faces are unordered; however, post-ranking via \textbf{N}o-\textbf{R}eference \textbf{I}mage \textbf{Q}uality \textbf{A}ssessment (\textbf{NRIQA}), \textbf{G}enerated \textbf{I}mage \textbf{Q}uality \textbf{A}ssessment (\textbf{GIQA}), and \textbf{F}ace \textbf{I}mage \textbf{Q}uality \textbf{A}ssessment (\textbf{FIQA}), the sequences diverge from human judgment, identifying more evidently forged images as high-quality. In contrast, our method, based on facial perception, effectively classifies fake images.

\section{Discussion on zero-shot learning and fine-tuning.}
Due to the specificity of the ranking labels, our quality assessment metric is better at predicting ranks than predicting scores. In real-world scenarios, we may be able to achieve alignment with the image quality distribution in a given scene by relying on a small amount of manual labeling. The trained model available, as shown in Table \ref{tab:fine_tuning} (labeled data from the available \textbf{D}eep\textbf{F}ake \textbf{G}ame \textbf{C}ompetition on \textbf{V}isual \textbf{R}ealism \textbf{A}ssessment (\textbf{DFGC-VRA}) training set labels).

\begin{table}[tb]
\centering
\caption{Zero-shot and fine-tuning for face swapping.}
\begin{tabular}{lcccc}
\hline
\multirow{2}{*}{Settings} & \multirow{2}{*}{Metric} & \multirow{2}{*}{Consistency $\uparrow$} & \multicolumn{2}{c}{Correlation $\uparrow$} \\ \cline{4-5}
 &  &  & SRCC & PLCC \\ \hline
\multirow{3}{*}{Zero-shot} & MUSIQ & 59.06 & \textbf{0.305} & 0.244 \\ 
 & SDD-FIQA & 54.24 & 0.252 & 0.203 \\ 
 & Our & \textbf{68.37} & 0.267 & \textbf{0.246} \\ 
\hline
\multirow{3}{*}{Fine-tuning} & MUSIQ & 70.22 & 0.647 & 0.633 \\
 & SDD-FIQA & 64.63 & 0.576 & 0.560 \\
 & Our & \textbf{90.11} & \textbf{0.849} & \textbf{0.832 }\\ \hline
\end{tabular}
\label{tab:fine_tuning}
\end{table}

\section{More results of improved face swapping.}
\begin{figure*}[h]
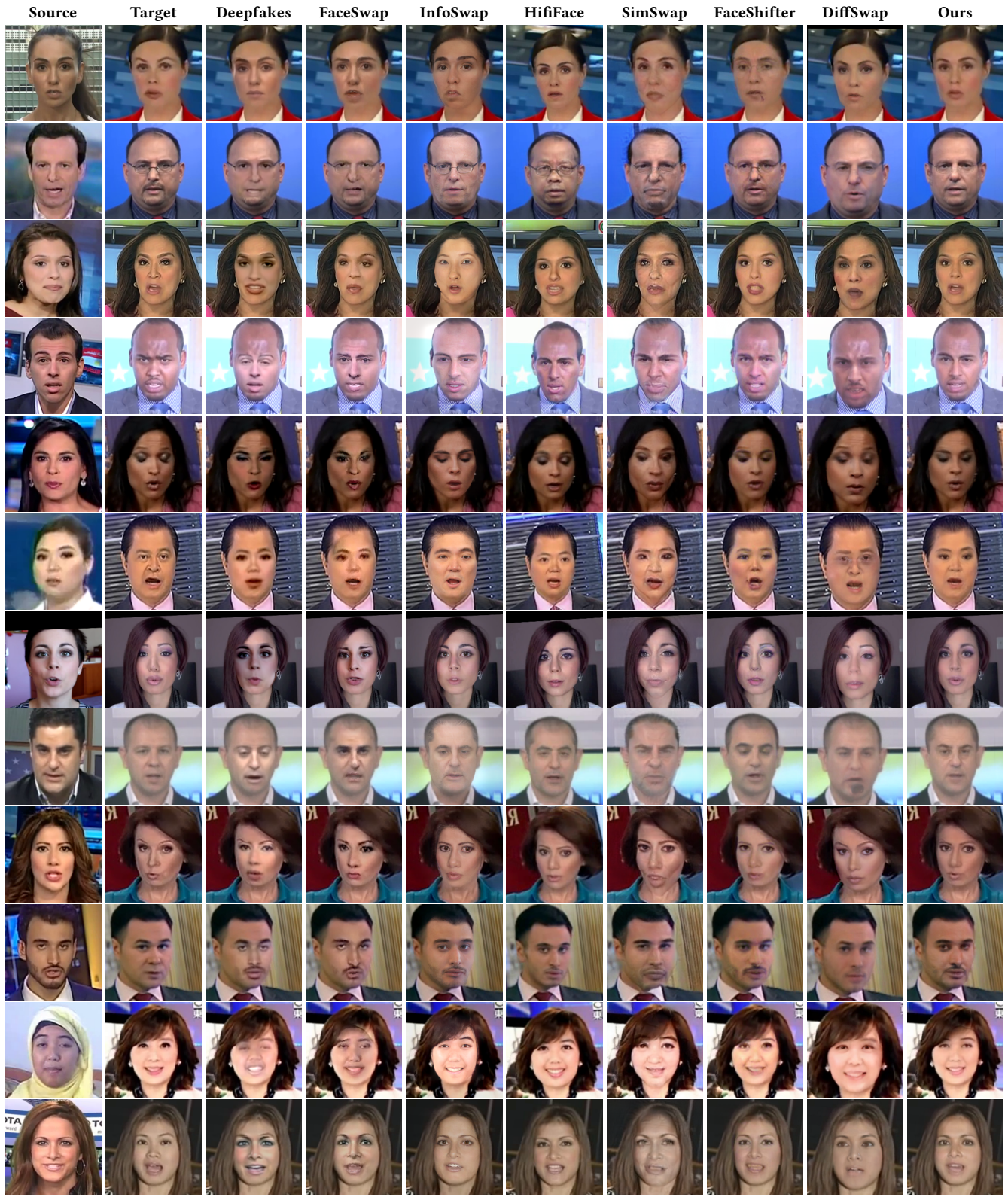

\centering
\foreach \row in {01,02,03,04,05,06,07,08,09,10,11,12}{
    \ifnum\row=01
        \foreach \col/\name in {0src/Source, 0tgt/Target, df/Deepfakes, fs/FaceSwap, is/InfoSwap, hf/HifiFace, sw/SimSwap, fsh/FaceShifter, diff/DiffSwap, ours/Ours}{
            \begin{subfigure}{0.095\textwidth}
                \centering
                \caption*{\textbf{\name}} 
                \includegraphics[width=\linewidth]{image_supp/candi_\row/\col.png}
            \end{subfigure}
        }
    \else
        \foreach \col in {0src,0tgt,df,fs,is,hf,sw,fsh,diff,ours}{
            \begin{subfigure}{0.095\textwidth}
                \centering
                \includegraphics[width=\linewidth]{image_supp/candi_\row/\col.png}
            \end{subfigure}
        }
    \fi
    \par
}
\caption{More qualitative comparison on FaceForensics++.}
\label{fig:more_results}
\end{figure*}

As shown in Figure \ref{fig:more_results}, we demonstrate more results that help improve the image quality of face swapping. Compared to other methods, the quality-improved face-swapping model is better at preserving the expressions and poses of the target faces.

\section{Ablation of quality metric.}
We added ablation experiments in relation to $L_{attr}$ and ablation experiments in the weights of $L_{quality}$. When ID retrieval is based on ArcFace, we achieve the best performance across all full-reference quality assessment metrics. When ID retrieval based on AdaFace, we can achieve significant performance improvements on facial attributes with imperceptibly minor errors, as shown in Table \ref{tab:ablations}. As shown in Table \ref{tab:weight_of_loss}, we demonstrate the trade-off between ID retrieval and facial attributes under different loss function weights, where we select a ratio of $40:1$ as the final proportion. In this setting, optimal results are achieved across all metrics.

\begin{table}[tb]
\centering
\caption{Ablation of loss in quality-improved face swapping.}
\begin{tabular}{lcccc}
\hline
\multirow{2}{*}{Settings} & \multicolumn{2}{c}{ID Retrieval $\uparrow$} & \multirow{2}{*}{Pose $\downarrow$} & \multirow{2}{*}{Exp $\downarrow$} \\ \cline{2-3}
 & ArcFace & AdaFace & & \\
\hline
w/o any attribute loss & 87.55 & 89.76 & 3.79 & 4.21 \\
w/ $L_{attribute}$ & 93.34 & \textbf{95.38} & 2.94 & 2.48 \\
w/ \textbf{$L_{quality}$} (\textbf{Ours}) & \textbf{95.01} & 95.25 & \textbf{2.57} & \textbf{2.32} \\
w/  $L_{quality} + L_{attribute}$ & 91.40 & 92.73 & 2.80 & 2.71 \\
\bottomrule
\end{tabular}
\label{tab:ablations}
\end{table}

\begin{table}[htbp]
\centering
\caption{Weights of quality loss function, referenced by ID. The best result is highlighted in boldface, and the second best result is highlighted in underline.}
\begin{tabular}{lccccc}
\toprule
\multirow{2}{*}{Setting}  & \multirow{2}{*}{Ratios} & \multicolumn{2}{c}{ID Retrieval $\uparrow$} & \multirow{2}{*}{Pose $\downarrow$} & \multirow{2}{*}{Exp $\downarrow$} \\ \cline{3-4}
 & & ArcFace & AdaFace & & \\
\midrule
w/o \( L_{\text{quality}} \) & & 87.55 & 89.76 & 3.79 & 4.21 \\ 
SOTA method & & - & - & 2.94 & 2.48 \\
\hline
\multirow{4}{*}{w/ \( L_{\text{id}} \) \& \( L_{\text{quality}} \)} & 10:1 & 86.98 & 88.60 & 2.31 & 2.24 \\
 & 20:1 & 91.77 & 92.44 & \textbf{2.37} & \textbf{2.29} \\
 & 40:1 & \textbf{95.01} & \textbf{95.25} & \underline{2.57} & \underline{2.32} \\
 & 80:1 & \underline{94.89} & \underline{95.12} & 2.93 & 2.47 \\ 
\bottomrule
\end{tabular}
\label{tab:weight_of_loss}
\end{table}